\documentclass[sigconf,anonymous=false]{acmart}
\def\BibTeX{{\rm B\kern-.05em{\sc i\kern-.025em b}\kern-.08emT\kern-.1667em\lower.7ex\hbox{E}\kern-.125emX}}
    
\definecolor{dkgreen}{rgb}{0,0.6,0}
\definecolor{gray}{rgb}{0.5,0.5,0.5}
\definecolor{mauve}{rgb}{0.58,0,0.82}

\usepackage{listings}
\usepackage{algorithm}
\usepackage{algorithmic}
\usepackage{multicol}

\usepackage{subcaption}

\usepackage{amsmath}
\usepackage{breqn}

\usepackage [english]{babel}
\usepackage [autostyle, english = american]{csquotes}
\MakeOuterQuote{"}

\def\code#1{\texttt{#1}}

\newtheorem{theorem}{Theorem}[section]

\newtheorem{myexample}[theorem]{Example}
\newtheorem{myremark}[theorem]{Remark}
\newtheorem{mydefinition}[theorem]{Definition}

\acmConference[A short version of this paper appears in NeSy'19 @ IJCAI 2019]{13th International Workshop on Neural-Symbolic Learning and Reasoning @ IJCAI 2019}{August 2019}{Macau, China}
\acmYear{2019}
\copyrightyear{2019}

\begin{document}

%
\title{One-shot Information Extraction from Document Images using Neuro-Deductive Program Synthesis}

%
\author{Vishal Sunder$^1$, Ashwin Srinivasan$^2$, Lovekesh Vig$^1$, Gautam Shroff$^1$, Rohit Rahul$^1$}
\affiliation{$^1$TCS Research, $^2$BITS Pilani, Goa}
\email{s.vishal3@tcs.com, ashwin@goa.bits-pilani.ac.in, lovekesh.vig@tcs.com, gautam.shroff@tcs.com, rohitrahul@tcs.com}
%
%
%
%

%
\renewcommand{\shortauthors}{Vishal, et al.}

%

\begin{abstract}
Our interest in this paper is in meeting a rapidly growing industrial demand for information extraction from images of documents such as invoices, bills, receipts etc. In practice users are able to provide a very small number of example images labeled with the information that needs to be extracted. We adopt a novel \lq two-level\rq `neuro-deductive', approach where (a) we use pre-trained deep neural networks to populate a relational database with facts about each  document-image; and (b) we use a form of deductive reasoning, related to meta-interpretive learning of transition systems to learn extraction programs: Given task-specific transitions defined using the entities and relations identified by the neural detectors and a small number of instances (usually 1, sometimes 2) of images and the desired outputs, a resource-bounded meta-interpreter constructs proofs for the instance(s) via logical deduction; a set of logic programs that extract each desired entity is easily synthesized from such proofs. In most cases a single training example together with a noisy-clone of itself suffices to learn a program-set that generalizes well on test documents, at which time the value of each entity is determined by a majority vote across its program-set. We demonstrate our two-level neuro-deductive approach on publicly available datasets ("Patent" and "Doctor's Bills") and also describe its use in a real-life industrial problem.
\end{abstract}

%
%

%

\maketitle

\section{Introduction}
Extraction of information from structured documents has long been an important problem in the research and application
of Information Retrieval (IR) techniques. A challenging version
of this task arises when the contents of the documents are not already
in a structured database, but are captured as images. In industrial
settings, this is especially common: images of invoices, bills, forms, etc.,
are readily available, and information needs to be extracted from them (``What is the address to which this invoice was sent?'', {\em etc}\/.). The images can be noisy (captured using a low-quality camera, at a sub-optimal angle, for example), and can even be ambiguous and the information extraction task is often manually done.

With the rapid advancement of Deep Learning (DL) for computer vision problems, many DL architectures are available today for document image understanding (\cite{konya2012adaptive}, \cite{oliveirafast}, \cite{raoui2017deep}, \cite{wang2017deep}). But like most DL-based techniques, training these models from scratch is resource and data
intensive. This is a major stumbling block for
industrial problems for which collecting and annotating data incur significant costs in time and money.
In this paper, we use two complementary forms learning to address
this problem:

\begin{enumerate}
    \item \textit{Neural-learning:} Using pre-trained DL models for reading document images and converting them into a structured form by populating a predefined database schema.
    \item \textit{Deductive-learning:} Using the entities and primitive relations identified by
        neural-learning, synthesize re-usable logic programs for extracting entities from
        a document image, using proofs constructed by a meta-interpreter in a manner similar to
        explanation-based generalization (EBG: \cite{ebg:sld}), and generalizing the
        proofs using techniques developed in Inductive Logic Programming (ILP: \cite{muggleton1994inductive}).
\end{enumerate}

\begin{figure}
\includegraphics[width=\linewidth]{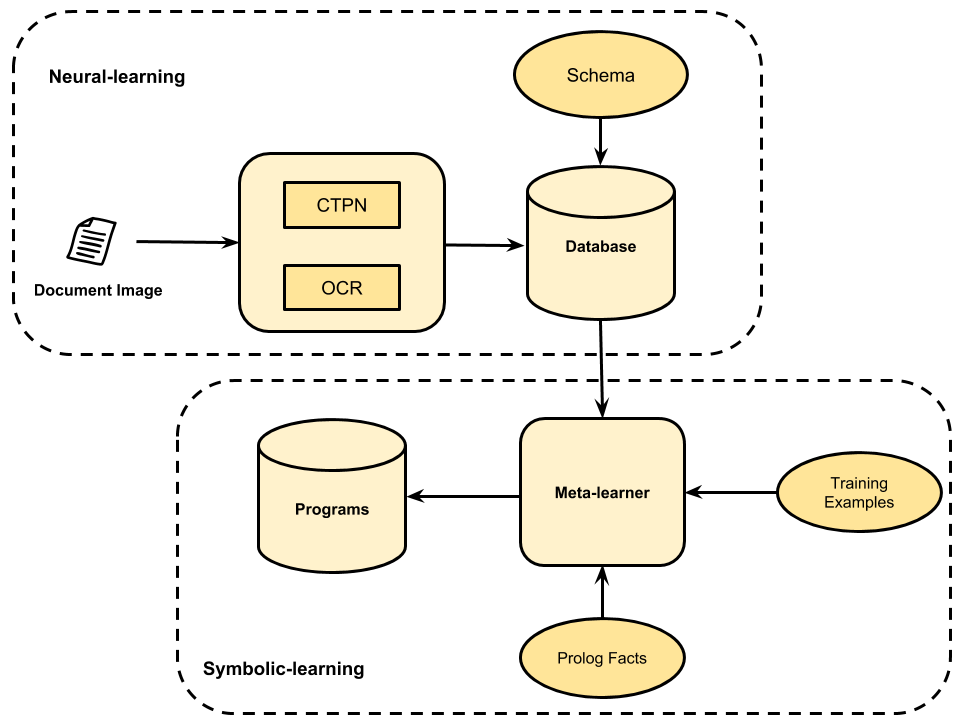}
\caption{Overview of the system\label{fig: block}}
\end{figure}

A schematic overview our approach is given in Fig. \ref{fig: block}.
The choice of logic-based EBG for symbolic learning has two attributes that are of
interest to us: (1) EBG methods generalize from a single
data instance (or sometimes just a few) by exploiting strong domain constraints (in
effect, the constraints act as a prior over possible models for the data);
and (2) The logical models can often be converted into an
(human-)interpretable form. This makes it possible to allow
human intervention, which is important in practice.

The neuro-symbolic learners are used to deploy a
one-shot learning strategy for information extraction from
images of documents of a particular kind (invoices, for example).
Given one training instance,
the neural-learning results in a database containing the objects and
relations recognised in the corresponding image. The symbolic
learner than synthesizes a (re-usable) program
that can extract entities from any document of the same kind. Some additional machinery is needed for tackling outlier cases. We allow human intervention to make corrections by providing few additional annotations (usually
just one or two). This is made possible by the interpretable nature of
the output produced by symbolic learning. Although such cases are few in number, they is an important step towards building robust
``human-in-the-loop'' systems
(\cite{xin2018accelerating}) that use human expertise to enhance their performance.


This paper makes the following contributions:

\begin{enumerate}
\item It combines deep-learning based image processing and deductive program synthesis to address industrial problems
involving information extraction from document images.
\item It reports on the implementation of an end-to-end one-shot
    learning strategy for synthesizing programs that extract entities from a document image.
\item It presents a human-in-the-loop based $N$-shot learning algorithm for extracting entities from document images in 
      when required due to outlier cases.
\end{enumerate}

\noindent
These contributions are supported by results on some publicly available
datasets. We also describe, within the constraints allowed, results on a small proprietary dataset which is nevertheless indicative of the industrial applications of the approach.

The rest of this paper is organized as follows: Section \ref{sec:vision} gives an overview of the DL-based vision stage. Section \ref{sec:meta-prog} formalizes the approach used for program synthesis. Section \ref{sec: noise} describes our one-shot learning technique.
It also proposes a human-in-the-loop based few-shot learning algorithm. Results are then presented and discussed in Section \ref{sec:results}. Section \ref{sec:analysis} provides a performance analysis of the results and a few ideas for enhancement and debugging. We conclude in Section \ref{sec:concl}.

\section{Neural Learning: Image $\rightarrow$ Database}
\label{sec:vision}
Given a document image, the task of extracting spatial relationships between different entities in the document and subsequently populating a database schema is handled by a suite of deep vision APIs which we shall refer to as the \textit{VisionAPI} in the rest of the paper.

The VisionAPI comprises of two modules: \textit{1) A Visual detection and recognition module 2) Spatial-relationship generation module.}

\subsection{Visual detection and recognition}
The job of this module is to locate the \textit{bounding boxes} around horizontally aligned text in an image. Post detection, recognition amounts to inferring the text present in this bounding box via OCR.

\subsubsection{Text Detection}
For this task, we use state-of-the-art Connectionist Text Proposal Network (CTPN) \cite{tian2016detecting} which is commonly used for text detection in scene text images. This is a Convolutional Neural Network (CNN) which takes an image as input and generates text for the given image. The sequence of proposals are then passed to a Recurrent Neural Network (RNN). This allows the network to exploit the contextual visual features of continuous text. The output is in the form of bounding box coordinates around the text. We use a pretrained version of this network (trained on the ICDAR 2013 dataset, \cite{karatzas2013icdar}).

\subsubsection{Recognition of Entities}
The bounding boxes returned by the text detection module are then cropped from the input images and each one of the box is then fed to an OCR module (Optical Character Recognition). We have used the Google Vision API for this purpose but in principle any other OCR like Tesseract can also be used. As a result , we get a string corresponding to the text that is inferred by the OCR module in the unicode format. For each word of the string, we apply a data-type detection module which identifies the abstract data-type of the word. (For example, $<name>,<city>,<date>,<word>,<alphanumeric>$ etc)

\subsection{Identification of Primitive Relations}
To generate basic spatial relationships, we exploit the bounding box coordinates and the corresponding text that we get from the previous module. In this manner, we generate $17$ different relationships each of which come under one of the following categories. (In principle, one can come up with any number of such relationships given the bounding box coordinates and the corresponding text.)
\begin{enumerate}
\item \textit{Text blocks:} We define a textblock as a set of lines which begin at approximately the same x-coordinate and the vertical distance between them is not more than twice the height of the text line (calculated through the coordinates). This yields  relationships which give us the words and lines that are part of the same text block. 
\item \textit{Page lines:} A line of text which is horizontally aligned is defined as a page line. We build a relationship which is a mapping between a word and the  page line in which it occurs. 
\item \textit{Above-below:} This captures the relationships between lines and blocks in the vertical direction. We have $4$ relationships in this category where for every line and every block, we have an above and below relationship indicating which lines/blocks are above or below other lines/blocks .
\item \textit{Left-right:} This category is the same as above-below except that it is in the horizontal direction and has additional relationships between words. 
\item \textit{Substring:} For a line/block, we have a relationship which maps every pair of words in that line/block to the substring between the pair. To account for multiple occurrences of word pairs, we assign a unique index to it. 
\item \textit{Datatype:} Similar to substring except that here, instead of word pairs we use datatype pairs. \end{enumerate}

\begin{table}
\centering
\begin{tabular}{c}
\toprule
\code{text\_blocks\_master}, 
\code{page\_lines\_master} \\
\code{lines\_below\_block\_word}, 
\code{word\_in\_line} \\
\code{above\_block}, 
\code{below\_block} \\
\code{above\_line}, 
\code{below\_line} \\
\code{word\_right\_left}, 
\code{right\_block} \\
\code{left\_block}, 
\code{right\_line} \\
\code{left\_line}, 
\code{block\_to\_substring} \\
\code{block\_to\_substring\_dtype}, 
\code{line\_to\_substring} \\
\code{line\_to\_substring\_dtype} \\
\bottomrule
\end{tabular}
\caption{Primitive relations obtained from the VisionAPI.\label{tab:pos_rel_list}
}
\end{table}

A tabulation of the relations found by the neural-learner is in Table \ref{tab:pos_rel_list}, which define the domain
theory for program synthesis. (A more detailed explanation of the deep-learning techniques used in the VisionAPI
can be found in \cite{rahul2018deep}.)

\section{Deduction: Database $\rightarrow$ Programs}
\label{sec:meta-prog}
The task of the deductive stage is to automatically construct a programmatic mechanism
to extract entities from the database populated by the neural learner, for all documents in the same template-class of
`similar' documents.

Deductive program synthesis can be seen as a form of Explanation-Based Generalization (EBG) \cite{ebg:sld}.
Although EBG was originally formulated for concept learning,
here the goal is to identify programs that implement functions.
Given a training example $e$ that identifies the output(s) $O$ from
some input image $I$, the task is to identify a program that is sufficient to
compute $O$ from $I$, given $B$. Conceptually, this is done in two steps: (a) A {\it proof\/}
is constructed for the computation of $O$ from $I$, given $B$; and (b) The proof is {\it generalised\/} to obtain a (re-usable) program.

In this paper, we adopt the representation of logic programs for $B$:
In our case $B$ consists of (Prolog) rules  defined using the primitive
relations in the database populated by 
the neural learning stage (\ref{tab:pos_rel_list}). 

\begin{myexample}
 Here are two background definitions (in Prolog syntax) for an ``invoice'' template, that
 use the primitive relations identified by the VisionAPI.
 \begin{center}
\begin{verbatim}
 has_keyword(Word,[In],[In,LineId,WordId]):-
    word_in_line(In,_,_,_,LineId,Word,WordId).
 
left_of(RWord,[In,LineId,RWordId],[In,LineId,LWordId]):-
  word_right_left(In,_,LWord,LType,LWordId,RWord,RType,RId),
  word_in_line(In,_,LType,_,LineId,LWord,LWordId),
  word_in_line(In,_,RType,_,LineId,RWord,RWordId).
 \end{verbatim}
 \end{center}
 The first rule definition looks for a word \code{Word} in the \code{word\_in\_line} relation and returns its \code{WordId} and the \code{LineId} of the line its contained in.
 The second rule definition returns the \code{LWordId} of the word to the left of \code{RWord}.
 \end{myexample}

For the computational system we use the operational semantics of a transition system (in the sense identified by Plotkin in \cite{plotkin}).
\begin{mydefinition}
{\bf (Transition system)}
A logic program defining a simple transition system $T$ is: \\
\begin{tabular}{ll} 
\hspace*{0.2cm} $ts((C,C)) \leftarrow$ \\
\hspace*{0.2cm} $ts((Ci,Cf)) \leftarrow$ \\
 \hspace*{0.4cm} $trans(T,Ci,C)$, \\
\hspace*{.4cm} $ts((C,Cf))$
\end{tabular}
\end{mydefinition}
\noindent
As defined above, the transition system can compute indefinitely, and in practice, we
impose a bound on the computation, by including a depth-limit.
The $C$'s are used to denote {\em configurations} that can be more general than states. With this definition a training example $v$ is
then a simply a specification of a specific input-output configuration pair $ts((i,o))$, where $i$ is a document, in which the
value of the desired entity $e$ is $o$.
\begin{myremark}
\label{def:ts}
{\bf (Transition Systems and Meta-interpretation)}
A depth-bounded version of the transition system, $T_d$ can be
readily implemented as a logic program (shown here as a Prolog
program):
\begin{center}
\begin{verbatim}
ts((Ci,Cf),D):-
        D = 1,
        trans(T,Ci,Cf).
ts((Ci,Cf),D):-
        D > 1,
        trans(T,Ci,C),
        D1 is D - 1,
        ts((C,Cf),D1).
\end{verbatim}
\end{center}
\end{myremark}

Given this definition, a depth-bound $d$ and a domain-theory $B$
consisting of definitions of $trans/3$ predicates, We will
say a depth-bounded explanation exists for $v = ts((i,o))$
if a Prolog interpreter is able to prove $e$ using SLD-resolution
and denote this by:
\[
    B \wedge T_d \vdash v
\]
In \cite{ebg:sld}, a simple modification of the usual Prolog interpreter is described that
allows a retention of the proof-tree. The modification employs
additional clauses $B_M$ for proving Prolog clauses, and:
\[
    (B_M \wedge B \wedge T_d \vdash  prove(v,P)) ~{\equiv}~(B \wedge T_d \vdash v)
\]
\noindent
Here, $prove(e,P)$ denotes ``$e$ can be proved using $P$'' . We will
call $B_M$ a meta-interpreter for clauses in $B \wedge T_d$, and
$P$ as the set of literals that are $TRUE$ in a meta-interpretive proof for $v$
(the elements of $P$ are obtained from the goals,
or negated literals, that resolve in a refutation-proof for $v$).

Programs are constructed by generalising the literals obtained in a meta-interpretive proof.

\begin{myexample}
A fragment of a document image is shown below:

\centerline{\includegraphics[width=\linewidth]{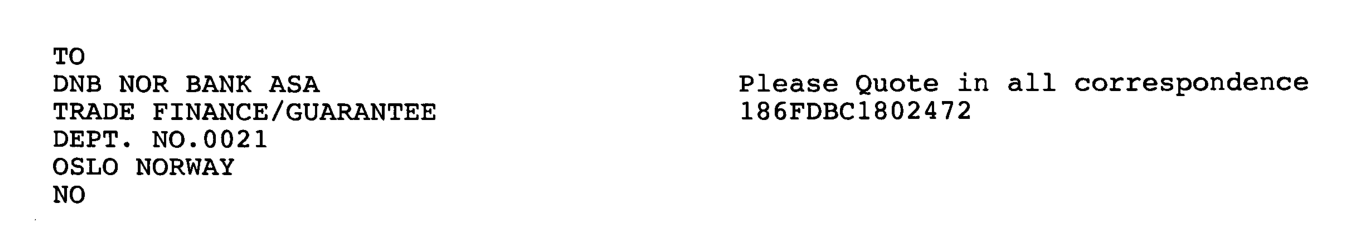}}
\end{myexample}

From such images we want to extract the reference  number for correspondence. 
The neural-learner extracts primitive entities and relations as Prolog facts.
The training instance provided to the program synthesis stage is
the input-output pair $([d1],[186FDBC1802472])$, where $d1$ is
the identifier of the image above. Program
synthesis then is the result of the following steps:
\begin{enumerate}
    \item A meta-interpretive proof $P$ for $ts(([d1],[186FDBC1802472])$
consists of the set of ground literals:\\
        \begin{tabular}{ll}
        $\{trans(has\_keyword('Please'),[d1],[d1,loc1]),$ \\
        $has\_line\_below([d1,loc1],[186FDBC1802472]\}$
        \end{tabular}
        
    \item The literals in $P$ are used construct a ground clause
        $G$ (this is the ``explanation'' step):
        
        \begin{tabular}{ll}
            $ts(([d1],[186FDBC1802472])) \leftarrow$\\
            \hspace{0.2cm} $has\_keyword('Please'),[d1],[d1,loc1])$ \\
            \hspace*{0.2cm} $has\_line\_below\_word([d1,loc1],[186FDBC1802472])$
            \end{tabular}
            
\item The ground clause is generalised and with some trivial
    renaming, resulting in the final program definition (this is
    the ``generalization'' step):
    \begin{verbatim}
        corr(A,B) :-
                has_keyword('Please', A,C),
                has_line_below(C,B).
    \end{verbatim}
\end{enumerate}

Given domain constraints $B$, there can be many proofs and corresponding
programs, for each training example $v$ for a desired entity $e$.
Meta-interpretive program synthesis, referred to from here on as $MIP(D,B,v)$, returns a set of programs to extract the value
of the entity $e$ from the database $D$ populated by the neural learner. 
As we shall argue and demonstrate below, the same set of programs can also be used to extract the value of
$e$ from other similar documents.

In general $MIP(D,B,v)$ produces a number of programs given a single training instance, thus raising the difficulty
of choice. One way to address that may reduce the number of possible
explanations is to provide more examples, that add more
constraints (we seek an explanation now for all the examples). Traditionally,
EBG has prefered the specification of some extra-logical criterion for selection
amongst multiple explanations. In an industrial setting, both of these options
translate to requiring high-cost expertise. We describe next a one-shot
learning with a form of ``re-sampling'' that is surprisingly effective.

\section{One-shot Learning: Noisy Cloning}
\label{sec: noise}

In general, providing more than one training instance should result in programs
that is in some sense ``more general''.
Since our task is to extract entities in any document in a template-class (and not just one document in the class), it is important that the program synthesized applies to as many documents as possible in the class (we will discuss outliers later).

\begin{myexample}
An unsatisfactory program for extracting the
number for correspondence is the one below:
\begin{center}
\begin{verbatim}
corr(A,B) :-
       has_keyword('Please',A,C),
       left_of('Please',C,D),
       right_of('ASA',D,C),
       has_line_below_word(C,B).
\end{verbatim}
\end{center}

The program has unnecessary conditions \code{left\_of} and \code{right\_of}.
It also uses the keyword 'ASA' which is part of the address to
the left of the correspondence number. The address will change in other documents rendering the program incorrect for these.
\end{myexample}

One way of correcting the result of program synthesis is to keep adding example-pairs until the programs become correct. This method is data intensive and one would need to annotate multiple documents of a given template. Instead, we create
additional examples automatically by ``noisy cloning'', that is shown
later to be surprisingly effective in practice.

A document $d'$ is a noisy clone of a document $d$, if: 

\begin{enumerate}
\item $d$ and $d'$ have the same template; and
\item $d$ and $d'$ have the same entities, but each entity in $d$ has a
    different value to the entity in $d'$
\end{enumerate}

Given a document $d$, we can obtain a noisy clone $d'$ by simply altering the values of all its entities. This can be seen as a form of re-sampling with noise added to all entity
values in $d$, resulting in two documents. Provided the
the entity to be extracted from $d$ occurs only once in $d$, this form
of re-sampling can assist in generalising programs identified using
a single example. We propose two different procedures
for program synthesis using \textit{few-shot} learning:
\subsubsection{TrainOS}

\begin{algorithm}[tb]
\caption{\textit{TrainOS($A$, $I_{train}$, $B$)}}
\label{alg:trainos}
\begin{flushleft}
\textbf{Given}: 
    (1) A training example $e$ consisting of a
        document image, $I_{train}$ and
        the corresponding annotation, $A$ for $m$ entities of that document;
    (2) A database $D$; and
    (3) Domain rules and facts $B$ defined based on the database schema.\\
\textbf{Find}: A set of programs $P$ for all the entities.
\end{flushleft}
\begin{algorithmic}[1] 
\STATE Populate the database $D = VisionAPI(I_{train})$
\FOR{all $(f_i,v_i^{train}) \in A$} \STATE Create a noisy clone, $\widetilde{v_i^{train}}$ \ENDFOR
\STATE Create a noisy copy, $\widetilde{D}$ for $D$ by replacing $v_i^{train}$ with $\widetilde{v_i^{train}}$ in $F$.
\FOR{$i:=1$ to $m$} \STATE Get a set of $n_i$ programs, \\ $\bigcup\limits_{j=1}^{n_i} p_{i}^{j} := MIP(D,B,v_i^{train}) \bigcap MIP(\widetilde{D},B,\widetilde{v_i^{train}})$ \ENDFOR
\STATE Return $P := \bigcup\limits_{i=1}^{m} \bigcup\limits_{j=1}^{n_i} p_{i}^{j}$
\end{algorithmic}
\end{algorithm}

Algorithm \ref{alg:trainos} corresponds to the \textit{TrainOS} algorithm (corresponds to Train One-Shot) which requires just one document image, $I_{train}$ and an annotation for all entities in that document, $\bigl\{(f_i,v_i^{train})\bigr\}_{i=1}^m$ to find programs for all entities ($m$ of them) in the corresponding document template. Here, $(f_i,v_i^{train})$ corresponds to the $i$th entity-value pair. Given this annotation, we add some noise to each of the entity values and create a noisy counterpart which is then used as a second "training" instance.

Thus, we get two training examples corresponding to the training document and its noisy counterpart. Completing the proofs for these two examples and then taking the intersection for the two sets of logical programs that follow is equivalent to finding programs which can extract entities from both the documents.

\begin{figure*}
\centering
\begin{subfigure}{.5\textwidth}
  \centering
  \includegraphics[width=0.8\linewidth]{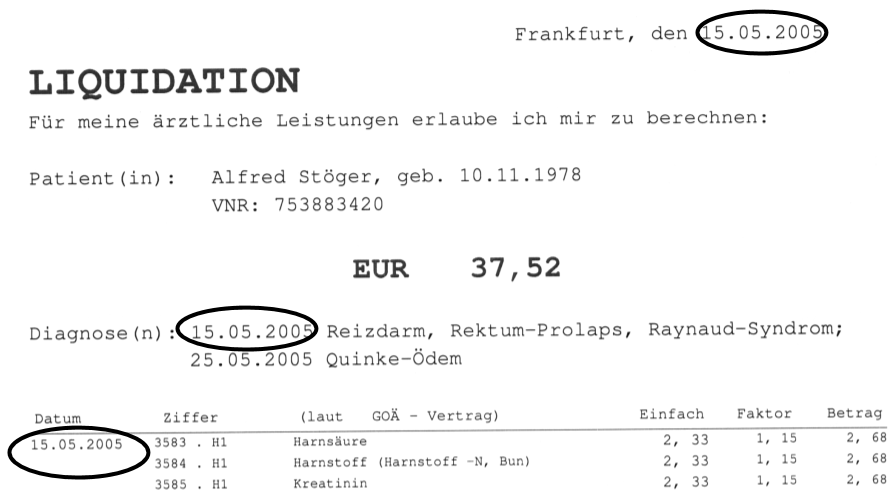}
  \caption{The $3$ dates marked in oval are same for this document.}
  \label{fig:loc_doc1}
\end{subfigure}%
\begin{subfigure}{.5\textwidth}
  \centering
  \includegraphics[width=0.8\linewidth]{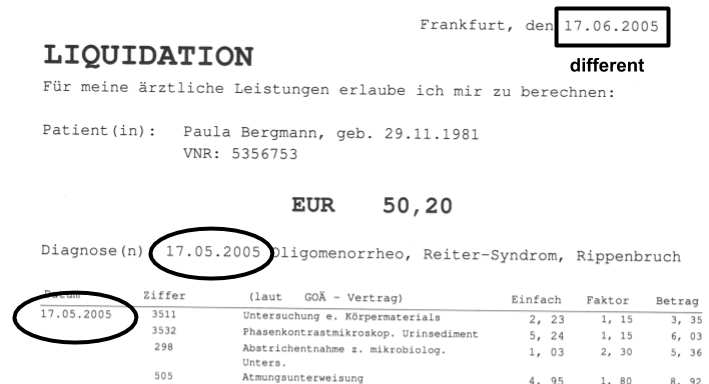}
  \caption{The $2$ dates marked in oval are the same but the one marked in the rectange is different.}
  \label{fig:loc_doc2}
\end{subfigure}%
\caption{Location ambiguity}
\label{fig:loc_ambi}
\end{figure*}

\subsubsection{TrainNS}
One of the instances where the \textit{TrainOS} fails to produce generalized programs are for entity values that occur at multiple locations in the document. In such scenarios, it becomes impossible for the system to disambiguate between these locations to get the actual position of the entity. The actual position of the entity may include all the positions where it occurs in the training document or just a subset of these positions. Figure \ref{fig:loc_ambi} shows an example of such a situation. Notice that in figure \ref{fig:loc_doc1} the date appears in three different locations. The actual position of the entity "date" is not evident from only this document. All three or less may be the actual location of this entity. But running the TrainOS algorithm would be assuming that the same entity always occurs at three different locations for any document of this template. 

The only way to clear this ambiguity is to provide another document with an annotation for the corresponding entity. Figure \ref{fig:loc_doc2} shows another example of the same template. This example clearly disambiguates between the three locations as now we are sure that the first location corresponds to a different entity. We are still not sure about the other two locations but giving more documents to the meta-interpreter will solve this issue.

We facilitate this by modifying the TrainOS algorithm to incorporate a "human-in-the-loop" approach, where a human is asked to annotate another document from a set, $l$ of supplementary documents ($I_{supp} = \bigcup\limits_{i=1}^{l} I_i$) whenever an entity occurs more than once. Therefore, until there is complete disambiguity in the entity locations, a human will be asked to feed in more annotated documents. Note that this annotation is done only for the one, ambiguous entity. If the human annotates $k$ different documents ($0 \leqslant k \leqslant l$) and these documents are also used for training (i.e. each document corresponds to an additional training instance), line $7$ of algorithm~\ref{alg:trainos} becomes:
\begin{multline}
\bigcup\limits_{j=1}^{n_i} p_{i}^{j} := (\bigcap\limits_{o=1}^k MIP(D_o,B,v_i^{o}) \bigcap MIP(D,B,v_i^{train}) \\ \bigcap MIP(\widetilde{D},B,\widetilde{v_i^{train}}))
\end{multline}
Here, each annotated document, $o$ will correspond to an additional document $D_o$.

We call this extension of TrainOS algorithm the \textit{TrainNS} algorithm (corresponds to Train N-Shot).

\begin{algorithm}[tb]
\caption{\textit{Extraction($I_{test}$, $P$, $f_i$)}}
\label{alg:extract}
\begin{flushleft}
\textbf{Input}: A document image, $I_{test}$, the set of all programs $P$ obtained for the corresponding template and the entity to be extracted $f_i$.\\
\textbf{Output}: The entity value $v_i^{test}$ corresponding to $f_i$.
\end{flushleft}
\begin{algorithmic}[1] 
\STATE Populate the database $D=VisionAPI(I_{test})$.
\FOR{all $p_{i}^{j} \in P$} \STATE $o_j := p_{i}^{j}(f_i)$ \ENDFOR
\STATE Get the set of distinct outputs, $\bigcup\limits_{j=1}^{K} o_k$ from $\bigcup\limits_{j=1}^{n_i} o_j$ sorted by their frequency of occurance (decreasing order). Here, $K \leqslant n_i$.
\IF{$o_1 = NULL$} \STATE $v_i^{test} := o_2$ \ELSE \STATE $v_i^{test} := o_1$ \ENDIF
\STATE Return $v_i^{test}$
\end{algorithmic}
\end{algorithm}

\subsection{Entity extraction}
Once we get the programs for all entities in a document template, we extract an entity from a new document (not used for training) using algorithm \ref{alg:extract}.

For the $i$th entity $f_i$ in a document template, we get $n_i$ different programs. We run each of these programs and store their outputs. After this, a \textit{majority voting} technique is used wherein we take the most frequently produced output as the correct output. If the most frequent output is "NULL" (which may happen in case a program does not return anything), then we consider the second most frequent output as correct.

\begin{table*}
\begin{center}
\begin{tabular}{|c|c|c|c|c|c|c|c|c|}
\hline
\multicolumn{3}{|c|}{\textbf{Doctor-1}} & \multicolumn{3}{|c|}{\textbf{Doctor-2}} & \multicolumn{3}{|c|}{\textbf{Patent}}\\
\hline
\textit{Entity name} & \textit{TrainOS} & \textit{TrainNS} & \textit{Entity name} & \textit{TrainOS} & \textit{TrainNS} & \textit{Entity name} & \textit{TrainOS} & \textit{TrainNS} \\
 \hline
 Date-1 & \textbf{91.00} & 91.00 (0) & Date-1 & 81.71 & \textbf{100.00} (1) & Classification No. & \textbf{77.67} & 77.67 (0) \\
 \hline
 Date-2 & \textbf{91.11} & 91.11 (0) & Date-2 & 58.28 & \textbf{90.00} (2) & Abstract & \textbf{68.50} & 68.50 (0) \\
 \hline
 Amount-1 & \textbf{85.33} & 85.33 (0) & Amount & 80.00 & \textbf{100.00} (1) & Applicants name & \textbf{86.67} & 86.67 (0) \\
 \hline
 Amount-2 & 60.40 & \textbf{100.00} (2) & Invoice no. & \textbf{100.00} & 100.00 (0) & Application No. & \textbf{77.50} & 77.50 (0) \\
 \hline 
 Patient name & \textbf{92.00} & 92.00 (0) & Patient name & \textbf{100.00} & 100.00 (0) & Representative & \textbf{74.33} & 74.33 (0) \\
 \hline
 Patient address & \textbf{97.78} & 97.78 (0) & Patient address & \textbf{100.00} & 100.00 (0) & Title & \textbf{78.25} & 78.25 (0) \\
 \hline
 Diagnosis & \textbf{100.00} & 100.00 (0) & Diagnosis & \textbf{79.43} & 79.43 (0) & Publication date & \textbf{81.30} & 81.30 (0) \\
 \hline
 Ref. no. & \textbf{100.00} & 100.00 (0) & VNR no. & \textbf{100} & 100 (0) & Inventors name & \textbf{83.33} & 83.33 (0) \\
 \hline
 NA & NA & NA & DOB & \textbf{100} & 100 (0) & Filing date & \textbf{70.00} & 70.00 (0) \\
 \hline
\end{tabular}
\caption{Entity extraction accuracy on 3 different public datasets (in \%) for \textit{TrainOS} and \textit{TrainNS} approaches for program generation. The number in brackets gives the number of supplementary documents (on average) required.\label{tab:results_retri_pub}}
\end{center}
\end{table*}

\section{Results}
\label{sec:results}
We use the following publically available datasets for testing our system and benchmarking results:
\begin{enumerate}
\item \textit{Doctor's Bills dataset:} This dataset is a collection of invoices for medical bills (\cite{van2008document}). It comprises of two different templates which we call Doctor-1 and Doctor-2 that have a total of 50 and 40 documents respectively. For each template, we keep aside 5 documents in the \textit{training pool} and the rest for testing. We do manual annotation for each of the templates and mark 8 entities for Doctor-1 and 9 entities for Doctor-2 which are to be extracted.
\item \textit{Patent dataset:} This dataset is a part of the \textit{Ghega dataset} \cite{medvet2011probabilistic}. It comprises of 136 patent forms from 10 different templates with annotations for different entities. For 8 of the templates, we keep 3 documents in the training pool and the rest for testing. We use 2 and 1 document(s) in the training pool respectively for two other templates as they contained a very few documents.
\end{enumerate}

\begin{figure*}
\centering
\begin{subfigure}{.333333\textwidth}
  \centering
  \includegraphics[width=0.8\linewidth]{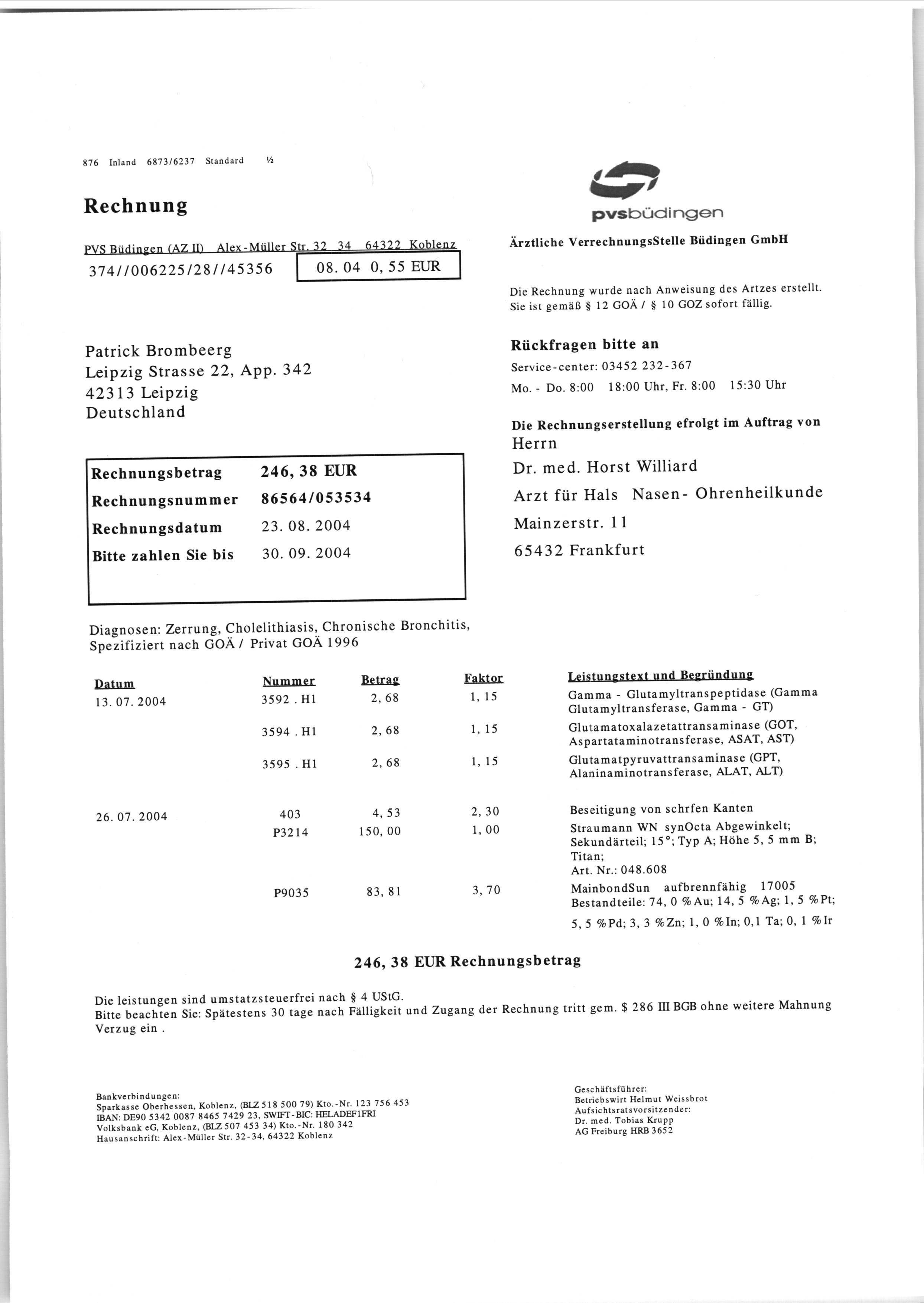}
  \caption{Doctor-1}
  \label{fig:doc1}
\end{subfigure}%
\begin{subfigure}{.333333\textwidth}
  \centering
  \includegraphics[width=0.8\linewidth]{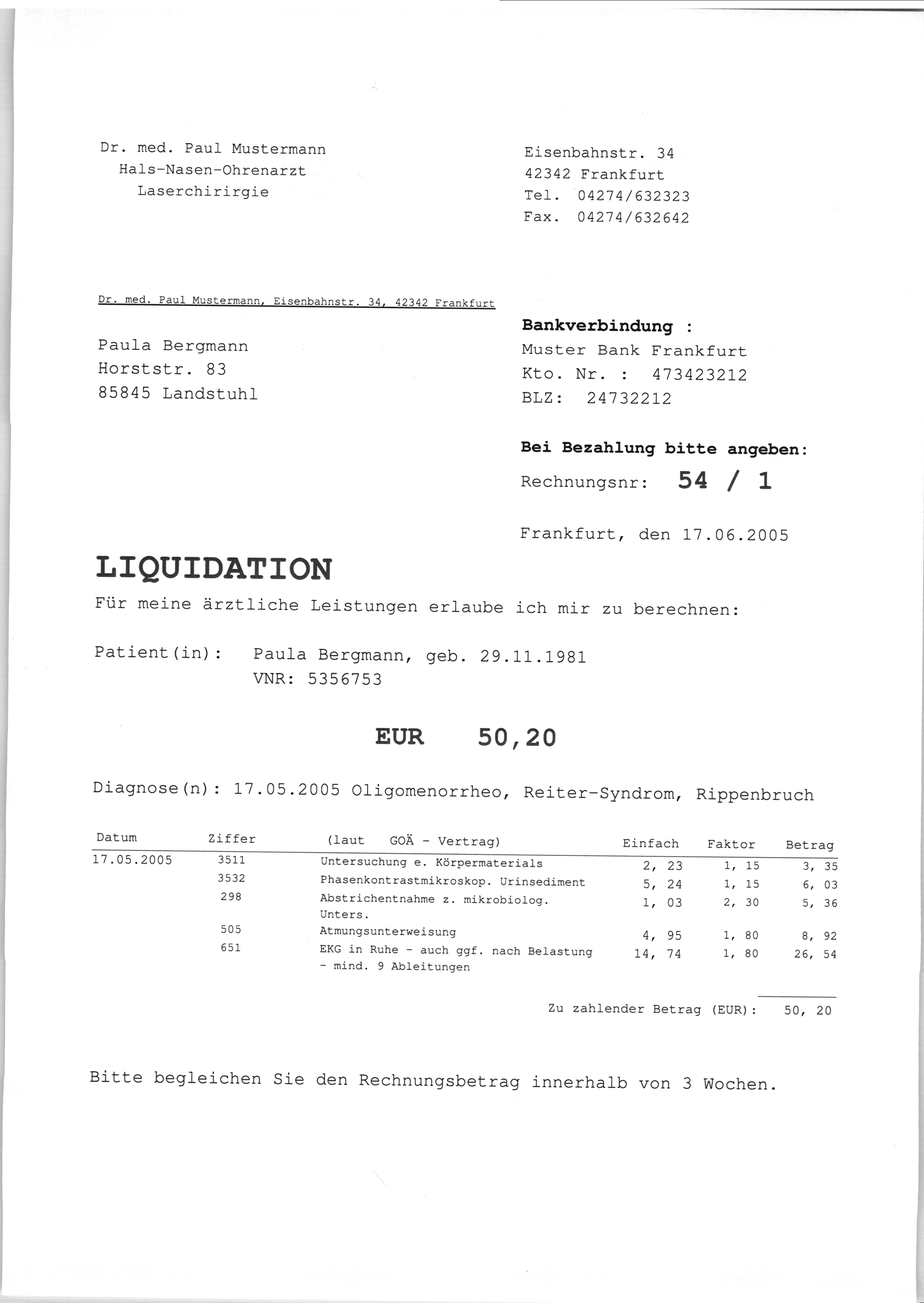}
  \caption{Doctor-2}
  \label{fig:doc2}
\end{subfigure}%
\begin{subfigure}{.333333\textwidth}
  \centering
  \includegraphics[width=0.8\linewidth]{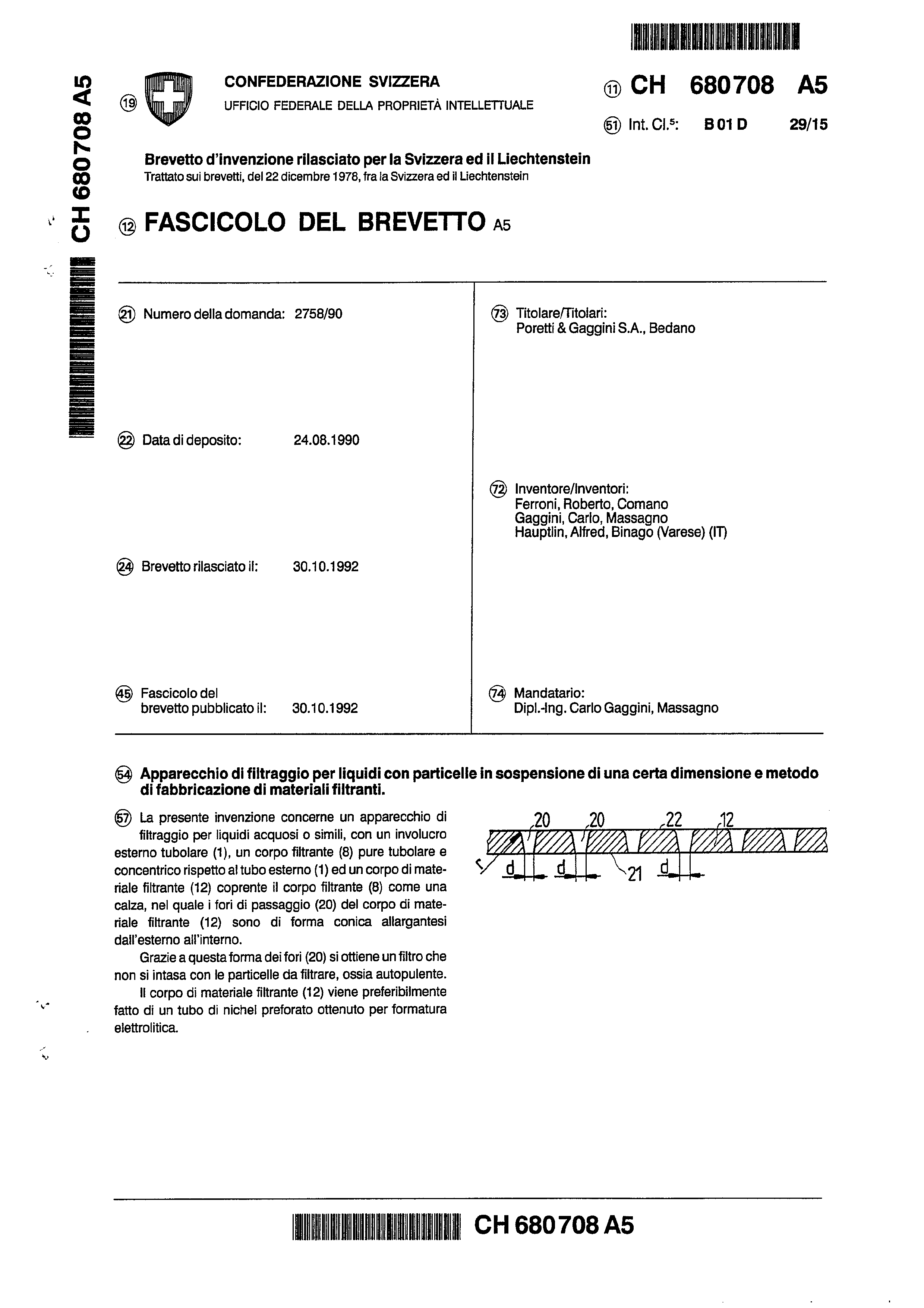}
  \caption{Patent}
  \label{fig:pat}
\end{subfigure}%
\caption{Sample document images}
\label{fig:sample_doc}
\end{figure*}

Figure \ref{fig:sample_doc} shows a document sample for each type.

We select a document from the training pool for one-shot learning and the rest for use as supplementary documents for running TrainNS algorithm. This is done multiple times such that each document in the training pool is used in one-shot learning once. The average extraction accuracy for every entity is reported in table \ref{tab:results_retri_pub}.

\begin{table}[H]
\begin{center}
\begin{tabular}{|c|c|}
\hline
\textit{Entity name} & \textit{TrainOS} \\
\hline
Account no. & 100.00 \\
\hline
Addressee & 100.00 \\
\hline
Amount & 100.00 \\
\hline
Contract no. & 100.00 \\
\hline
Correspondence no. & 100.00 \\
\hline
Drawee & 100.00 \\
\hline
Drawer & 100.00 \\
\hline
Tenor & 67.67 \\
\hline
\end{tabular}
\caption{Extraction accuracy (in \%) on proprietary data.\label{tab:results_retri_prop}
}
\end{center}
\end{table}

From the results, it is clear that for most entities (entities for which the TrainNS algorithm requires $0$ supplementary documents to be annotated), TrainOS algorithm is sufficient to obtain good extraction accuracies. For the cases where TrainNS algorithm gives better results, notice that the difference between extraction accuracies of TrainNS and TrainOS quite high (as high as 40\%). This is evidencial of the fact that location ambiguity is indeed a significant problem and the TrainNS algorithm alleviates this to a large extent. Also, note that even for such cases, the human had to annotate no more that $2$ documents from a supplementary document pool size of $4$ which cannot be considered a big overhead.

Although a good amount of work has been done to solve the problem of information extraction from document images, results by most of them are on proprietary documents which makes it impossible to test our method against a baseline. To the best of our knowledge, there is only one benchmarking results available on the Patent dataset by Ref.~\cite{medvet2011probabilistic} who use a probabilistic approach to solve the problem. Their task is slightly different from ours in that they detect the bounding box sequence for a given entity whereas we extract the exact value of the entity. Using one document, they report the average success-rate to be \textbf{50\%} whereas our one-shot method obtains an average accuracy of \textbf{77\%}. (However, with $14$ documents though, they achieve a success-rate of 90\% in their task of bounding-box detection).

Table \ref{tab:prog_eg} gives three examples of programs generated for Doctor-2 template and their interpretations. Each step of a program is completely interpretable unlike the intermediate steps (layers) in a deep neural network. This makes debugging of the system fairly simple which is of prime importance in any real-world deployment.
 
As an evidence of real-world application of our system, we also give results on one proprietary document template in table \ref{tab:results_retri_prop}. For this, we were given just one document for training and testing was done on three other documents of the same template. Note that we do not give results for the TrainNS approach as we had just one document for training.

\begin{table*}
\begin{center}
\begin{tabular}{clc}
\hline
\multicolumn{2}{|c|}{\textit{Program}} & \multicolumn{1}{|c|}{\textit{Interpretation}} \\
\hline
\multicolumn{1}{|r}{\code{date1(A,B):-}} & \code{get\_blockid(\lq Diagnosen\rq,A,C),} & \multicolumn{1}{|l|}{C = The block with the word "Diagnosen"}\\
\multicolumn{1}{|r}{} & \code{get\_block\_above(0,C,D),} & \multicolumn{1}{|l|}{D = The block above C with index 0}\\
\multicolumn{1}{|r}{} & \code{get\_substring(\lq Rechnungsdatum\rq,\lq Bitte\rq,1,D,B).} & \multicolumn{1}{|l|}{B = Output (String in D between the two words in args)}\\
\hline
\multicolumn{1}{|r}{\code{diagnosis(A,B):-}} & \code{get\_line(\lq Diagnosen\rq,A,C),} & \multicolumn{1}{|l|}{C = The line with the word "Diagnosen"}\\
\multicolumn{1}{|r}{} & \code{get\_keyword(\lq <medical\_term>\rq,C,B).} & \multicolumn{1}{|l|}{B = Output (String in C which is a medical term)}\\
\hline
\multicolumn{1}{|r}{\code{ref(A,B):-}} & \code{word\_to\_left(\lq Rechnungsdatum\rq,A,B).} & \multicolumn{1}{|l|}{B = Output (Word to the left of \lq Rechnungsdatum\rq)}\\
\hline
\end{tabular}
\caption{Examples of programs generated by the meta-interpreter.\label{tab:prog_eg}}
\end{center}
\end{table*}

\begin{figure}
\includegraphics[width=\linewidth]{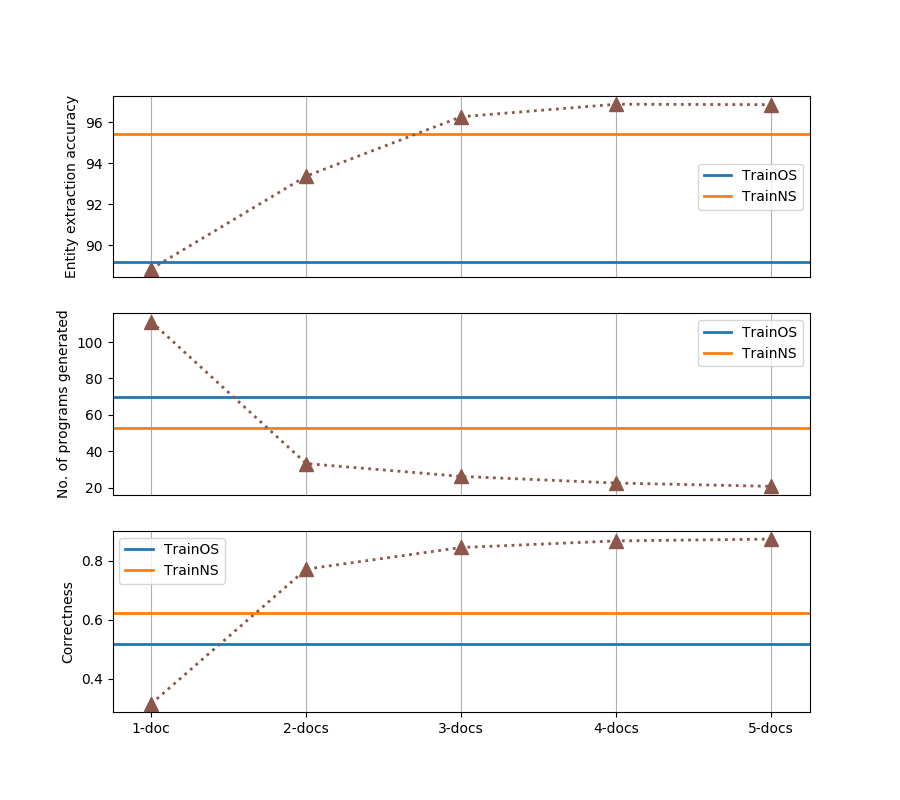}
\caption{Effect of increasing training size on performance.}
\label{fig:mult_doc_train}
\end{figure}

\section{Performance Analysis}
\label{sec:analysis}
We do a two-part analysis of our system performance. The first part pertains to the effect of the number of training documents on three different performance metrics. In the second part, an error analysis is done.
\subsection{Effect of training size}
As explained in section \ref{sec: noise}, the meta-interpreter has a potential to produce more generalised programs if more training examples are used to generate the programs. In other words, if the number of training documents are increased, we can expect the performance to get better. Thus, we experimented with this idea by providing the meta-interpreter with as high as 5 documents for training on the doctor's bills dataset (as we found that this has more documents per template compared to the patent dataset) and evaluated its performance on three metrics. We used all combinations of $n$ documents ($ 1 \leq n \leq 5$) and the average of their performance on the test set is shown in figure \ref{fig:mult_doc_train}.
\begin{enumerate}
\item \textit{Extraction accuracy:} This is the percentage of times the correct output is extracted. The extraction follows algorithm \ref{alg:extract}. The first plot of figure \ref{fig:mult_doc_train} shows this performance and it is clear from this plot that providing more training examples indeed makes the performance better. Also note that the extraction accuracy of the TrainNS algorithm is very close in performance to the best possible performance.
\item \textit{Number of programs generated and correctness:} Correctness is the fraction of programs that give the correct output and its variation with training size is given in the third plot of figure \ref{fig:mult_doc_train}. As we increase the training size, the number of programs generated by the meta-interpreter tends to fall rapidly as shown by the second plot of figure \ref{fig:mult_doc_train}. This is a consequence of the fact that as the training size increases, the meta-interpreter produces only the most general programs and hence the correctness score also increases.
\end{enumerate}
Note from the plot that both the TrainOS and TrainNS algorithms produce some non-general programs (their correctness values being close to 0.5 and 0.6 respectively). This we observed is because there are almost always some differences in the documents even within the same template. These differences may either be due to some noisy images or due to some minor formatting variations in the documents which effect the output of the VisionAPI. Introducing noise in such cases is not a complete solution for non-generality as the observation mentioned in section \ref{sec: noise} holds true for documents of the same template.

We observed that the extraction accuracy and the correctness curves of figure \ref{fig:mult_doc_train} reach a plateau as the number of training documents increase and in fact for some combination of the $5$ documents, there is a dip in these scores. This, we observe is because as the number of documents increase, there is a high probability that at least one of the documents is noisy. When this happens, it becomes impossible for the meta-interpreter to come up with programs that work for both the noisy and the non-noisy training instances.

Also note that there is a vast improvement in the correctness score when using just one document for training versus using TrainOS algorithm. This suggests that one-shot learning by using a noisy clone does improve performance (an absolute improvement as high as 0.5 on the correctness score) by narrowing down the output to the most general programs.
\subsection{Error analysis}
\subsubsection{Source of errors}
We observe that most of the errors in the extraction of an entity by one-shot learning is due to one of the following reasons:
\begin{enumerate}
\item \textit{Ambiguity in entity location:} These errors are mostly tackled when we use supplementary documents and run the TrainNS algorithm. This is evidenced in table \ref{tab:results_retri_pub} for cases where TrainNS accuracy is higher than TrainOS.
\item \textit{Inconsistencies in the output of the VisionAPI:}  As discussed before, these errors occur when a test image is either noisy or there are some formatting differences in the training and testing images. In such cases, we essentially have different templates during train and test time. Such errors have to be fixed by making the VisionAPI more robust which can be an interesting direction for future work.
\end{enumerate}

\begin{figure}
\includegraphics[scale=0.35]{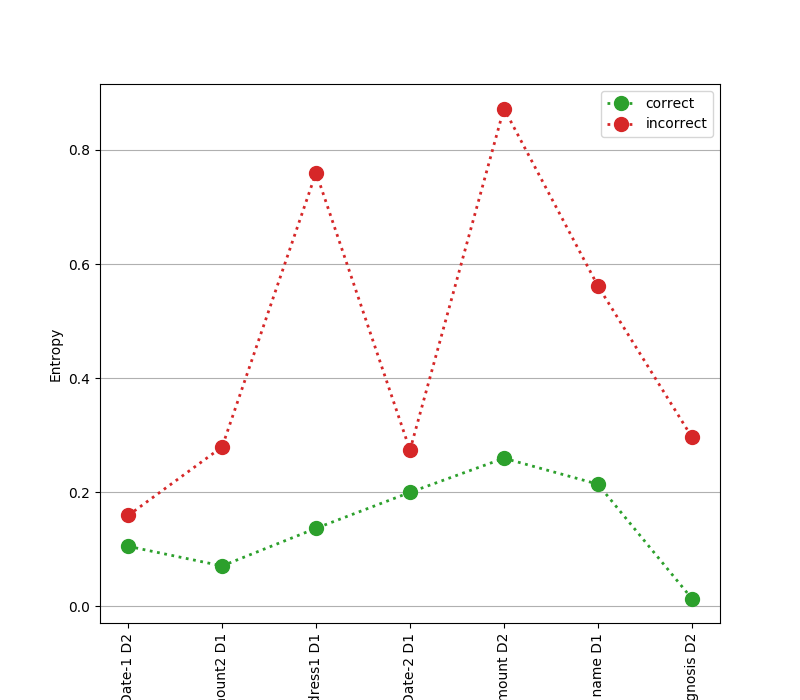}
\caption{Variation of entropy of different output distributions for correct and incorrect extractions.}
\label{fig:entropy}
\end{figure}

\subsubsection{Erroneous outputs are identifiable}
Any deployable system needs to be such that its debugging does not produce overheads. Hence, it is of prime importance that a failure instance is identifiable. In our case, these instances fall into one of the two categories: \textit{(1) No output extracted (2) Wrong output extracted}. From a dubugging point of view, errors that fall into category (1) are more desirable than (2). But we find the even the instances that produce a wrong output are identifiable to some extent.

For this, we calculate the entropy of the distribution over all distinct outputs produced by the programs for the cases when the extraction is either correct or incorrect but not for the cases when there is no extraction. This is done for every entity in the doctor's dataset and we show the plot for this in figure \ref{fig:entropy}. From this, we can clearly see that for correct output cases, the entropy values are always lower suggesting a more sparse distribution over the distinct outputs. Also, for every entity there is a visible difference between entropies of correct and incorrect predictions which suggests that using a suitable threshold for the entropy, incorrect extractions can be identified.



\section{Related Work}
Information Extraction from documents is a well established and explored field and a large body of work is available for this domain. But most of these works have focussed on structured or semi-structured documents rather than on document images. One such system is the RAPIER system \cite{mooney1999relational} which induces a pattern-match for entities using relational learning. This method is similar to ILP but is more constrained in terms of the rules it generates as it uses a slot-filling framework. On the other hand, our method works on document images and is diverse as it generates programs which, in principle, may use any number of spatial relationships. Another similar system is WHISK \cite{soderland1999learning} which learns regular expression like rules to extract patterns and like RAPIER, it works on semi-structured text rather than on document images. Effective extraction techniques have been proposed using shallow domain knowledge like in Ref.~\cite{ciravegna2001adaptive}. They introduce the $(LP)^2$ system which uses some domain knowledge to insert SGML tags on relevant entities. It then uses a correction mechanism to fine-tune the results. In constrast, our proposed method does not require any domain knowledge except the information in the documents themselves.

Efficiently extracting information from document images has been of some interest mainly for industrial applications. Ref.~\cite{ishitani2001model} have developed a system by graphically modeling relationships between words in the form of document models. They also use a predefined set of keywords and document models. Another approach particularly for templatized documents has been of learning a spatial structure from one part of a document and using this to extract entities from some other part \cite{bart2010information}. A method similar to ours, that uses a single document example to learn about spatial relations is proposed in Ref.~\cite{rusinol2013field}. This method represents word relations using a continuous polar coordinate system and rely on these for indexing different entities. This makes it sensitive to minor spatial variations. Another such real-world system is \textit{Intellix} \cite{schuster2013intellix}. This system uses categorization of entities based on position and context to come up with rules for extraction. Most of these works, to the best of our knowledge, do not benchmark results on public datasets which makes it difficult to validate results.

A reader familiar with Inductive Logic Programming (ILP) will find
a relationship of our program synthesis approach to the techniques used there. This is not surprising,
and EBG was shown to be a special case of ILP in \cite{muggleton1994inductive} given a single example.
The relationship to ILP extends, albeit more tenuously, to recent work on meta-interpretive learning (MIL) in that field. In MIL, proofs constructed by
a meta-interpreter using meta-rules in a higher-order logic are used
to instantiate first-order logic programs. Computation of outputs from inputs
can be seen as repeated applications of one or more
meta-rules, instantiated appropriately with predicates and terms from $B$.
Both standard ILP and MIL systems are more powerful (and complex to use)
than what is needed for our purpose.

Logic programming is used by Ref.~\cite{adrian2017document} to characterize the geometric properties of text units to reason over relationships between parts of a document. This approach is used in document understanding context. MIL (\cite{muggleton2014meta},\cite{muggleton2015meta}) has regained popularity in the recent past primarily because of its high interpretability and data efficiency. MIL has even proved to be much more robust as compared to many deep learning methods for certain applications in vision domain (\cite{dai2015logical},\cite{dai2017logical}). A particularly popular recent successful application of MIL in the one-shot learning context has been reported in Ref.~\cite{lin2014bias} where the authors replicate results of Microsoft's Flashfill by using one-shot MIL. 

ILP for Information Extraction has also been explored in the past by Ref.~\cite{ramakrishnan2007using} who have use ILP for extracting useful features. These features are then used to train machine learning models like SVM. It has been shown that ILP can come up with features that have exploitable signals to be used by the SVM. Another work which uses ILP for IE (\cite{junker1999learning}) defines only three primitive predicates and uses a seperate and conquer strategy to come up with rules for extraction. There is an underlying assumption in this approach that a document is a list of words. This makes it unsuitable in domains where two dimensional relationships between words are important as in templatized documents like invoices and forms. Our system incorporates relations like "above", "below" etc. and uses meta-interpretive program synthesis which makes it novel.

Our work extends the work in Ref.~\cite{bain2018identification} in the following ways. First, it shows that meta-interpretive learning of transition systems can be used beyond the biological domain used in that paper. Further, unlike in \cite{bain2018identification}, we do not allow invention of new transitions, and our transition system is not probabilistic.

\section{Conclusion}
\label{sec:concl}
In this paper, we presented a new approach for synthesizing programs that extract information from document images. To the best of our knowledge, this is the first attempt in combining Deep Learning based computer vision APIs with a Meta-Interpretive Learning based framework. Our approach is highly data efficient and in most cases requires just one document image to generalize well while requiring no more than three for a few cases. This is evidenced by results given on publically available datasets which can serve as a benchmark for future work. One-shot learning is highly indispensible in industrial scenarios where there is a limit on the proprietary documents that one can use for training modern deep learning based models. We have also shown that with the introduction of a few documents, our system can reach near perfect performance which is a direct consequence of using an Inductive Logic Programming (ILP) based approach for program synthesis.

Furthermore, after thorough analysis of the results, we conclude that there is still a lot of scope for improvement particularly in making the vision API more tolerant to noise and formatting variations in documents. Another direction worth pursuing in this context is to synthesize programs that are template invariant such that a part of the program is a template identifier. This would of course require incorporation of certain template specific rules in our current framework.
Even with this minor limitation, our system is highly robust and scalable to large-scale industrial applications. Finally, we have also begun applying the
same approach for very different document types, such as engineering drawings.

%
\bibliographystyle{ACM-Reference-Format}
\bibliography{sample-sigconf}

\clearpage
\appendix
\section{Supplementary Material}

\subsection{Computer Vision Components of the VisionAPI}
Besides the identification of the textual entities mentioned in section 3, we also perform the following additional visual pre-processing. 

\subsubsection{Image Alignment}
To correct for documents that were imperfectly aligned, we detect the bounding box around all of the high intensity pixels in the image and correct for any angular shift

\subsubsection{Image De-Noising}
We also address the issue of degradation in quality of images due to camera shake, improper focus, imaging noise, coffee stains, wrinkles, low resolution, poor lighting, or reflections. These kind of problems drastically affect the performance of many computer vision algorithms like text detection, OCR and localization. The objective here is to reconstruct high-quality images directly from noisy inputs and also to preserve the highly structured data in the images. We do this via an impementation of cylic GANs, details of which may be found in \cite{rahul2018deep}.  


\subsubsection{VisionAPISchema}
Once all the entities are identified as mentioned in section 3, relations between the entities need to be populated and stored in the database. So a schema should be designed to facilitate information extraction. All the entities are associated with their spatial coordinates and this information conveys the whereabouts of the neighbouring  text entities. This information is then  used to infer different logical and spatial relationships.

\begin{figure}[H]
\centering 
\includegraphics[scale=0.3]{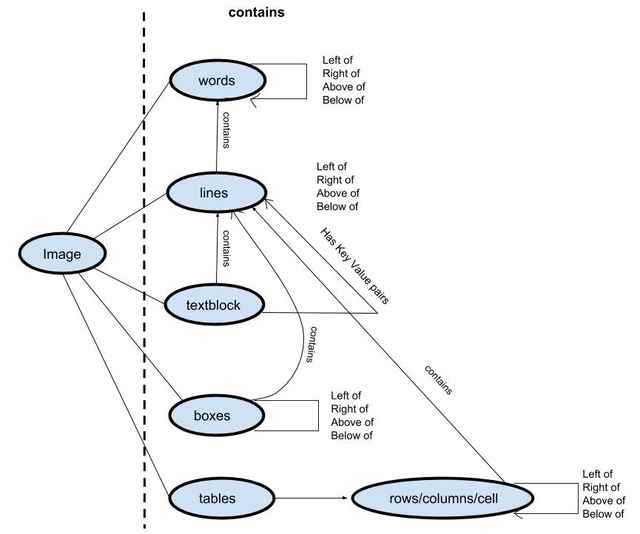}
\caption{VisionAPI Schema}
\label{schema}
\end{figure}

Figure \ref{schema} shows the representation of this schema populated in the database after the relevant relationships have been extracted from the raw image. The main entities of the schema includes words, lines, text blocks, boxes and tables. The inter and intra entity relationships have been illustrated by the directions of the arrow. The schema may get richer over time, we have only highlighted the entities that are useful for scanned document images at the moment. 

\subsection{Correctness of Program Synthesis}
Given the definition $T_d$ of a transition system as in Remark~\ref{def:ts};
a set of clauses $B$; and ground terms $i,o$. Let 
$t_1, t_2,\ldots,t_k$ denote $trans/3$ literals  in a meta-interpretive
proof for $ts((i,o))$.
For simplicity, we will assume that the $t_i$ are ground
(in general, they may contain existentially-quantified variables,
and a Skolemisation step will be needed for what follows).
Then, from the properties of the meta-interpreter and SLD-resolution we know
$B \wedge T_d \models (t_1 \wedge t_2 \wedge \cdots \wedge t_k)$.
Let $G: ts((i,o)) \leftarrow t_1, t_2, \ldots, t_k$ be
a ground clause. Then from the semantics of modus ponens,
it follows that $B \wedge G \models ts((i,o))$.
$G$ is called a ground explanation for $ts(i,o))$.
Let $H$ be a clause that $\theta$-subsumes $G$. It
follows from the properties of $\theta$-subsumption that
$B \wedge H \models ts((i,o))$. $H$ is
called a generalized explanation for $ts(i,o))$.

\subsection{Extract of Prolog code for the deductive reasoning module}
\begin{lstlisting}
MIP(D):-  % finds all possible programs with depth D 
	clean_up,
	set(cwa,true), %flag - ignore
	example(pos,Name,_,[Si,Sf]), % training example
	MIP_do((Si,Sf),Name,D),
	fail.
MIP(_,_).

MIP_do(S,Name,DepthBound):-
	S = (Input,Output),
	cts(S,DepthBound,Trace), 
	% prove transition for example
	trace_to_func(S,Name,Trace,Func), % generalization step
	check_soundness(Name,Input,Func), 
	% check that program derives other examples
	check_completeness(Name,_,Func), 
	% check that program derives nothing else
	update_cache(Func). % store the program in a cache

% depth-bounded transition system
cts((S,S),D,[]):- D >= 0.
cts((Si,Sf),D,[trans(T,Si,S)|Rest]):-
	D >= 1,
	D1 is D - 1,
	transition(T),			% domain transition
	trans(T,Si,S),
	cts((S,Sf),D1,Rest).

\end{lstlisting}

\end{document}